\documentclass{article}

    \PassOptionsToPackage{numbers, compress}{natbib}

 \usepackage[main, final]{neurips_2026}

\usepackage[utf8]{inputenc} 
\usepackage[T1]{fontenc}    
\usepackage{hyperref}       
\usepackage{url}            
\usepackage{caption}
\usepackage{mathtools}
\usepackage{booktabs}       
\usepackage{amsfonts}       
\usepackage{nicefrac}       
\usepackage{microtype}      
\usepackage{xcolor}         
\usepackage[utf8]{inputenc} 
\usepackage[T1]{fontenc}    
\usepackage{graphicx} 
\usepackage{hyperref}       
\usepackage{url}            
\usepackage{booktabs}       
\usepackage{amsfonts}       
\usepackage{amsmath}
\usepackage{nicefrac}       
\usepackage{microtype}      
\usepackage{xcolor}         
\usepackage{natbib}
\usepackage{amsfonts}       
\usepackage{amssymb}        
\usepackage{amsmath}
\usepackage{wrapfig}
\usepackage{cleveref}
\usepackage{enumitem}
\setlist{nolistsep}
\usepackage{multirow}
\setlist{nosep}
\usepackage{amsthm}
\theoremstyle{plain}

\theoremstyle{definition}
\usepackage{multirow}

\theoremstyle{remark}

\usepackage[table, dvipsnames]{xcolor}  
\usepackage[most,skins,theorems]{tcolorbox}
\usepackage{soul}
\newcommand{\pitheta}{\pi_\theta}
\newcommand{\y}{\mathbf y}
\newcommand{\z}{\mathbf z}
\newcommand{\ztheta}{\mathbf z_{\theta}}
\newcommand{\x}{\mathbf x}

\usepackage{needspace}

\tcbset{
  aibox/.style={
    width=\linewidth,
    top=7pt,
    bottom=2pt,
    left=2pt,
    right=2pt,
    colback=blue!6!white,
    colframe=black,
    colbacktitle=black,
    enhanced,
    center,
    attach boxed title to top left={yshift=-0.1in,xshift=0.1in},
    boxed title style={boxrule=0pt,colframe=white,},
  }
}

\newtcolorbox{AIbox}[2][]{aibox,title=#2,label=#1}

\NewDocumentCommand{\lifan}
{ mO{} }{\textcolor{cyan}{\textsuperscript{\textit{lifan}}\textsf{\textbf{\small[#1]}}}}
\NewDocumentCommand{\sagnik}
{ mO{} }{\textcolor{pink}{\textsuperscript{\textit{sagnik}}\textsf{\textbf{\small[#1]}}}}
\NewDocumentCommand{\hao}
{ mO{} }{\textcolor{purple}{\textsuperscript{\textit{hao}}\textsf{\textbf{\small[#1]}}}}
\NewDocumentCommand{\dilek}
{ mO{} }{\textcolor{blue}{\textsuperscript{\textit{dilek}}\textsf{\textbf{\small[#1]}}}}

\title{On the Primacy Bias in RLVR Training}

\author{%
  Phuc Minh Nguyen$^{1}$ \quad Ngoc-Hieu Nguyen$^{2}$ \quad Parshin Shojaee$^{3}$ \\\\
  \textbf{Chinh Duc La}$^{7}$ \quad \textbf{Duy Minh Ho Nguyen}$^{4,5,6}$ \quad \textbf{Rui Zhang}$^{2}$ \\ \\
  \textbf{Khoa D. Doan}$^{7}$ \quad \textbf{Ting Hua}$^{1}$ \quad \textbf{Nitesh V. Chawla}$^{1}$ \\ \\ 
  $^{1}$University of Notre Dame \quad
  $^{2}$The Pennsylvania State University \quad
  $^{3}$Virginia Tech \\ \quad $^{4}$University of Stuttgart\quad $^{5}$German Research Center for
Artificial Intelligence \\ $^6$Max Planck Research School for Intelligent Systems \quad $^7$VinUniversity\\
}

\begin{document}

\maketitle

\begin{abstract}
\looseness=-1 Reinforcement Learning with Verifiable Reward (RLVR) has emerged as a promising approach for enhancing the reasoning capabilities of large language models. However, it still remains unclear whether RLVR can extend reasoning capabilities beyond what is already encoded in the base model. 
In this paper, we study this question through the lens of \emph{primacy bias} — the tendency of models to over-rely on already-learned representations, disproportionately improving performance on problems they already handle well while underexploring harder ones that require departing from those representations. Through our controlled experiments and large-scale analysis, we show that RLVR training on mixed data tends to further sharpen performance on easier problems while yielding only marginal gains on harder ones. We find that although hard problems lead to modest gains overall, they are actually the ones that help most in expanding the model’s reasoning capabilities and pushing the representations. However, prolonged training on these hard problems may also hurt and lead to forgetting on easy problems.
These findings help to better understand some of the prior conflicting observations about RLVR, specifically, why on (OOD) domains where the base model performs poorly, improvements on hard problems can expand reasoning; while on well-trained domains, forgetting can reduce performance and mask genuine gains obtained from hard problems.

Code~\includegraphics[height=0.95em]{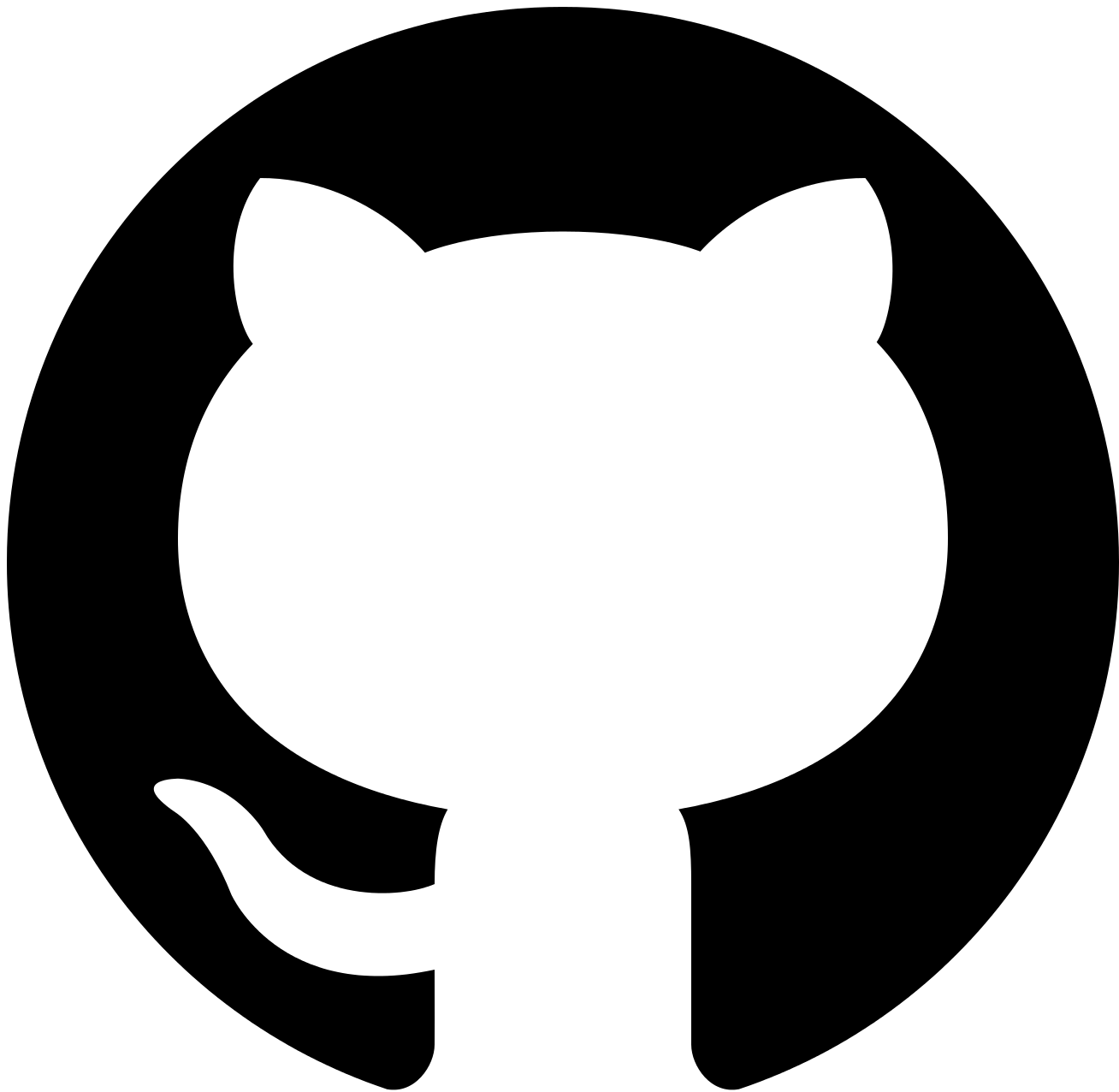}: \url{https://github.com/minhphucng30-UND/Primacy-bias-RLVR}
\end{abstract}
\section{Introduction}

\looseness=-1 Large language models (LLMs) fine-tuned for reasoning have recently achieved remarkable gains on complex reasoning benchmarks. A key driver is Reinforcement Learning with Verifiable Reward (RLVR) \citep{DeepSeekAI2025DeepSeekR1IR,lambert2025tulu}, which iteratively optimizes LLMs using simple, verifiable rewards, thereby incentivizing complex reasoning behaviors such as backtracking and self-verification within the reasoning traces of these models. 


\looseness=-1 Despite this progress, the answer to the question ``\textit{does RLVR expand toward new reasoning capabilities?}'' still remains under debate. Some studies have suggested that RL fine-tuning only improves sample efficiency on problems already solvable within the base model, rather than enabling it to solve new problems \citep{yue2025does,liu2025understanding,wu2026the}. On the other hand, some other works~\citep{research2026composer,liu2025prorl,sun2026rl, yuan2026from} have provided evidence that RL-trained models can still induce novel reasoning strategies on out-of-distribution (OOD) domains. These works, however, suggest that the relationship between the base model and the evaluated task---specifically the extent to which the base model can already solve the task problems---may play an important role in explaining whether RLVR leads to reasoning expansion or sharpening.

\looseness=-1 In this paper, we seek a principled explanation for these seemingly contradictory outcomes through the lens of \emph{primacy bias}---the tendency to overfit to initial experiences (i.e., the pre-trained knowledge) that can damage the rest of the learning process~\citep{kielo2024its, achille2018critical}. This bias not only limits exploration but also impairs the \emph{learnability} of on-policy data during RL training. It arises because the model becomes increasingly rigid and remains confined near its pre-trained initialization, rather than allowing meaningful adaptation of its internal representations towards new downstream tasks—a behavior commonly referred to as the \emph{lazy training regime} \citep{Lyle2025WhatCG, Chizat2018OnLT, Lyle2025WhatCG}.
\looseness=-1 To study this hypothesis, we first investigate primacy bias in RLVR using experiments that fully control both pre-training and post-training pipeline (\textbf{Section}~\ref{sec:primacy_bias_mnist}). We compare coverage dynamics during RL fine-tuning induced by pre-training against random initialization. Our results show that primacy bias leads to coverage shrinkage, regardless of the fine-tuning distribution. Furthermore, we observe that this primacy bias induces sparser parameter updates during training which hinders the network's ability to learn rare or more challenging problems (i.e., those that the pre-trained or base model struggles to handle). These observations suggest that the RLVR model mostly operates in the \textit{lazy regime}~\citep{Chizat2018OnLT,NEURIPS2018_5a4be1fa}, staying close to its pre-trained initialization and requiring prolonged training to meaningfully adapt to harder examples \citep{Neyshabur2014InSO,kumar2024grokking,Lyle2025WhatCG}. 

\looseness=-1 We then extend our analysis to LLM reasoning and investigate the existence of lazy training under RLVR (\textbf{Section}~\ref{sec:linearized}). Our results show that RLVR-trained models could often be well approximated by the linearization around their pre-trained base model, while exhibiting minimal feature drift. This result suggests that RLVR in this more general setting also operates in the lazy regime, providing a possible explanation for the sparse parameter updates observed during RLVR training, as reported in \cite{mukherjee2025reinforcement}.
We believe that these findings place RLVR within a theoretically tractable regime, which can provide a foundation for more principled, scalable, and analyzable studies of learning dynamics in reasoning post-training  \citep{Neyshabur2014InSO,kumar2024grokking,Canatar2020SpectralBA}.

\looseness=-1 Motivated by our findings, we identify an \textit{unlearnable phenomenon} that emerges from the interaction between primacy bias and the lazy training regime in RLVR-based LLM training (\textbf{Section}~\ref{sec:unlearnable}). To study this, we first perform fine-tuning across a mixture of easy and hard reasoning problems and observe that mixed training disproportionately improves rewards on easy examples while limiting learning on harder ones.
Our results show that although hard problems lead to only modest overall gains during training, they are actually the ones that help most in expanding the model’s reasoning capabilities.
At the same time, focused training on hard problems introduces a subtle trade-off that must be carefully taken into account.
We find that prolonged training on these difficult examples can also hurt and lead to catastrophic \textit{forgetting} on the easier problems, likely due to substantial deviations from the initial base model representations.
This explains why, as observed in \cite{liu2025prorl}, prolonged RLVR training can expand the reasoning boundary in less familiar or out-of-distribution domains, where the pre-trained model initially struggles. In such settings, extended training enables the model to gradually escape the lazy regime and meaningfully adapt to the new target tasks~\citep{kumar2024grokking, sun2026rl, yuan2026from}, ultimately leading to broader reasoning coverage.
In contrast, in domains where the pre-trained model already performs well—such as mathematical or code reasoning, particularly on easier problems—prolonged training may provide limited benefit and can even lead to adverse effects such as coverage shrinkage or catastrophic forgetting. In these cases, the gains obtained from adapting to harder examples may be outweighed by degradation on the easier ones due to substantial deviations from the base model representations and the effects of catastrophic forgetting.

We hope that our study provides a clearer understanding of the conflicting observations surrounding RLVR post-training and motivates more principled approaches for studying and controlling the trade-off between adaptation and forgetting.

In summary, our main contributions are as follows:
\begin{itemize}[leftmargin=*]
\looseness=-1 \item We unveil the primacy bias and lazy learning as the underlying mechanism to reconcile conflicting findings of whether RLVR expands or sharpens the reasoning boundary, where models tend to rely on base model's features when adapting to post-training data, limiting their ability to acquire new capabilities during RL fine-tuning.
  
\looseness=-1 \item  Through our controlled and large-scale empirical analysis, we demonstrate that RLVR operates in a near-linearized regime around the base model, with the learned representation remaining largely unchanged throughout the RL fine-tuning process. As a result, this leads models to disproportionately improve on easy problems while limiting their ability to effectively learn on hard ones.

\looseness=-1 \item Despite limited overall improvement, learning on hard problems can expand the reasoning boundary, but comes at the cost of forgetting on easy problems. This explains why, on OOD tasks, where forgetting is minimal, improvements on unsolved problems can increase Pass@$k$, whereas on over-trained domains, severe forgetting masks out improvement on hard problems.
\end{itemize}

\looseness=-1 \section{Related Works}
\textbf{RL Generalization of Reasoning LMs.} The role of reinforcement learning in driving generalization beyond the base model has been a central topic of study. Early works argue that RL-trained models may not acquire fundamentally new reasoning capabilities beyond those already present in their base models \citep{yu2025dapo, liu2025understanding, wu2026the}. In particular, 
\cite{yue2025does} demonstrates that RL training can even reduce the number of solvable problems, as measured by the Pass@$k$ metric, suggesting that performance is largely constrained by the base model’s capacity. Additionally, \cite{wu2026the} proposes a theoretical framework indicating that RL is unlikely to surpass the inherent limitations of the base model. By contrast, \cite{liu2025prorl} shows that extended RL training can enhance coverage on synthesized reasoning tasks. \cite{sun2026rl} further shows that RL can exhibit grokking and develop new reasoning strategies on extremely hard tasks. In contrast, we argue that these conflicting findings may arise from primacy bias: early learned features can hinder learning on hard mathematical tasks, while in OOD settings, where such biases are \emph{misaligned} with target tasks, RL can promote new reasoning strategies.

\looseness=-1 \textbf{Primacy Bias in Neural Networks.} Similar to biological systems, neural networks also exhibit critical periods—early phases of training during which representations remain highly plastic and are disproportionately shaped by initial data \cite{achille2018critical}. This phenomenon has later been explored in reinforcement learning, where it is often referred to as primacy bias \cite{kielo2024its}. Primacy bias occurs when the RL learning process is disrupted by overfitting to earlier experiences. One effect induced by primacy bias is the reduction of update/gradient norm, causing parameters to stop changing and hindering learnability during RL training \citep{pmlr-v232-abbas23a,kielo2024its}.
Recently, \cite{Lyle2025WhatCG} attributed primacy bias to limited feature learning, indicating a model's confinement to the lazy regime in which it fits the data without substantially updating its representations. This effectively behaves as a linear model, and consequently generalizes poorly. We build on this line by connecting coverage shrinkage to primacy bias, showing that primacy bias also arises in RLVR training, hindering learnability on hard problems.

\looseness=-1 \textbf{Neural Tangent Kernel.} The infinite-width limit is a well-studied model for deep network optimization \cite{NEURIPS2018_5a4be1fa}. In this regime, models remain close to initialization and behave like linear models governed by the initial Neural Tangent Kernel (NTK), which remains constant during training—known as the \emph{lazy training regime} \citep{Chizat2018OnLT}. Several studies have investigated the inductive biases in \emph{lazy training}\citep{Neyshabur2014InSO,Canatar2020SpectralBA,Bordelon2020SpectrumDL}, revealing that the NTKs preferentially fit data using the top eigenmodes of their spectrum, a phenomenon termed \emph{spectral bias}. While this
spectral bias can enable good generalization on some learning problems, target functions $\y(\x)$ that are not in the top eigenspaces of the kernel (misaligned), which can be measured by centered-kernel alignment (CKA) \citep{Cortes2012AlgorithmsFL}, require a potentially much larger amount of data. In contrast, in a rich (feature-learning) regime, the model can deviate substantially from its initialization, enabling meaningful changes to its internal representations, improving sample efficiency on target functions that are challenging for the lazy regime.\citep{Ghorbani2019LimitationsOL,kumar2024grokking,Arnaboldi2023FromH}. Furthermore, \cite{kumar2024grokking} characterizes the effect of delayed generalization (grokking) as a phase transition from lazy to rich (feature-learning) learning dynamics. In this paper, we show that RLVR models operate in the lazy regime, where RL-trained models remain close to the pre-trained model, inducing a bias toward solutions near the initial representation. As a result, optimization tends to prioritize improvements on easy problems while making little to no progress on harder ones, leading to a form of delayed generalization.
\section{Preliminaries}
\looseness=-1  \textbf{Language Model (LM).} We consider a language model $\pi_\theta$ parameterized by $\theta$, where the models learn to map an input prompt $\x$ to predictions $\y=\{\y_1,\cdots,\y_L\}\in\mathcal V^L$, where $\mathcal V$ is the vocabulary size, in an auto-regressive manner:
\begin{align}
    \pitheta(\y|\x) = \prod_{k=1}^{|\y|} \pitheta(\y_k|\x,\y_{<k}) = \prod_{k=1}^{|\y|}\text{Softmax}\left(\ztheta(\x,\y_{<k})\right)_{\y_k}
 \end{align} 
where $\y_k$ is the token at timestep $k$, and $\y_{<k}$ denotes the first $k-1$ tokens of $\y$, and $\ztheta(\x,\y_{<k}) \in\mathbb R^{\mathcal V}$ denotes the logits at the $k$-th token, and $|\y|$ is the response length.

\looseness=-1 \textbf{Coverage.} Coverage is the primary metric to quantify the reasoning capacity of LLMs. Let $r(\x,\y) \in \{0,1\}$ be a binary verifiable reward, indicating whether the response $\y$ is correct for a given problem $\x$. To measure the coverage of a language model $\pitheta$, the Pass@$k$ metric has been often used. This metric quantifies the probability that $\pitheta$ correctly solves a problem $\x$ within $k$ attempts, and is defined as:
\begin{align}
    \text{Pass@}k(\x,\pitheta) = 1 - (1 - \text{Pass@1}(\x,\pitheta))^k
\end{align}
\looseness=-1 where $\text{Pass}@1(\x,\pitheta) = \mathbb E_{\y\sim\pitheta(\cdot|\x)}\left[r(\x,\y)=1\right]$. In practice, Pass@$k$ is evaluated over a problem set $\mathcal{D}$ and measures the fraction of problems solved at least once by $\pi_\theta$ within $k$ attempts.


\vspace{-0.5em}
\section{Primacy bias and Lazy learning}
\label{sec:primacy_bias_mnist}
\vspace{-0.5em}

\looseness=-1 We study how pre-training–induced bias affects coverage expansion (or shrinkage) during RL training through the lens of primacy bias \citep{kielo2024its,achille2018critical}, a phenomenon in which early learning experiences hinder later learning phases. 
For example, pre-trained models tend to learn representations dominated by the majority classes, leaving rare classes underrepresented. 
Thus, RL finetuning is "initially restricted" by these "suboptimal" representations. 
Furthermore, we'll show that, under some conditions, the feature updates during RL stage are limited, a phenomenon known as lazy learning \citep{Lyle2025WhatCG, Chizat2018OnLT}. 
As a result, due to the primacy bias, hard (or rare) examples remain underexplored during the RL phase.
We hypothesize that this phenomenon consequently affects the \emph{reasoning boundary (coverage)} in LLM post-training. 



\begin{figure}[!h]
    \centering
    \includegraphics[width=0.95\linewidth]{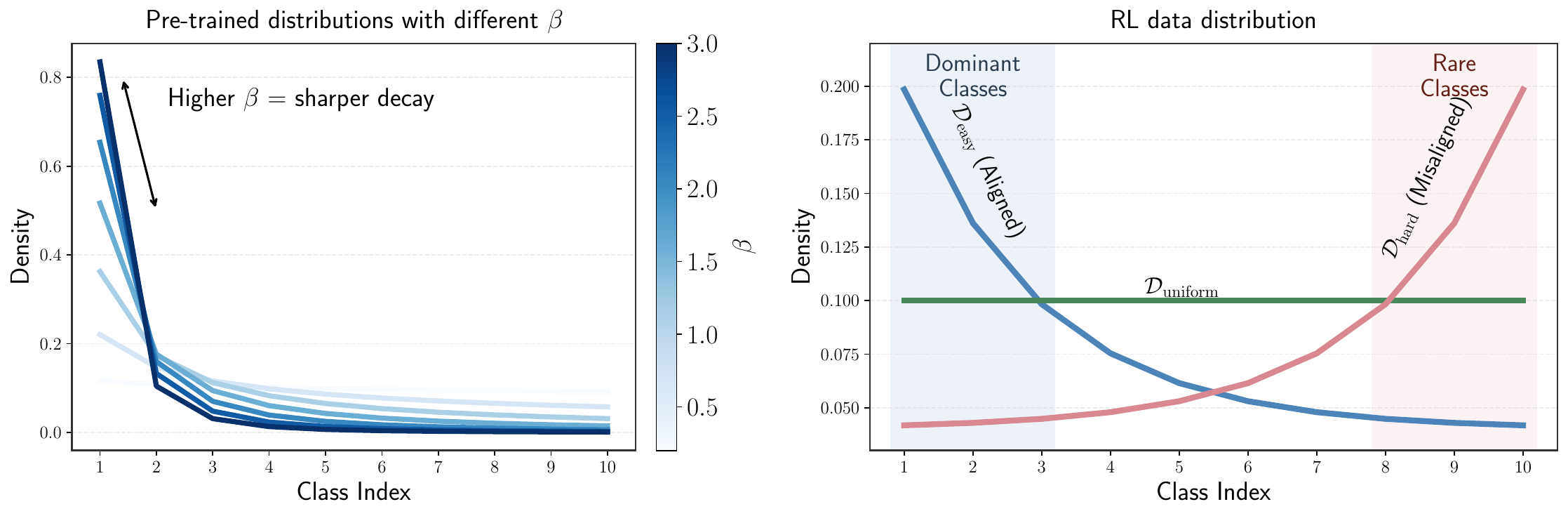}
    \vspace{-5pt}
    \caption{\looseness=-1 Experimental designs on toy example of image classification. 1)~The pre-trained data distribution is controlled by the skewness parameter $\beta$, greater skewness induces strong bias towards dominant classes. 2)~The class distribution of RL training data consists of $\mathcal{D}_{\text{easy}}$, $\mathcal{D}_{\text{hard}}$, and $\mathcal{D}_{\text{uniform}}$. Easy examples align with pre-trained knowledge, whereas hard examples are \emph{misaligned} with it and predominantly belong to rare classes.} 
    \vspace{-15pt}
    \label{fig:mnist_demonstration}
\end{figure}
\subsection{Primacy bias leads to coverage shrinkage} 
\begin{figure}[t]
    \centering
    \includegraphics[width=0.95\linewidth]{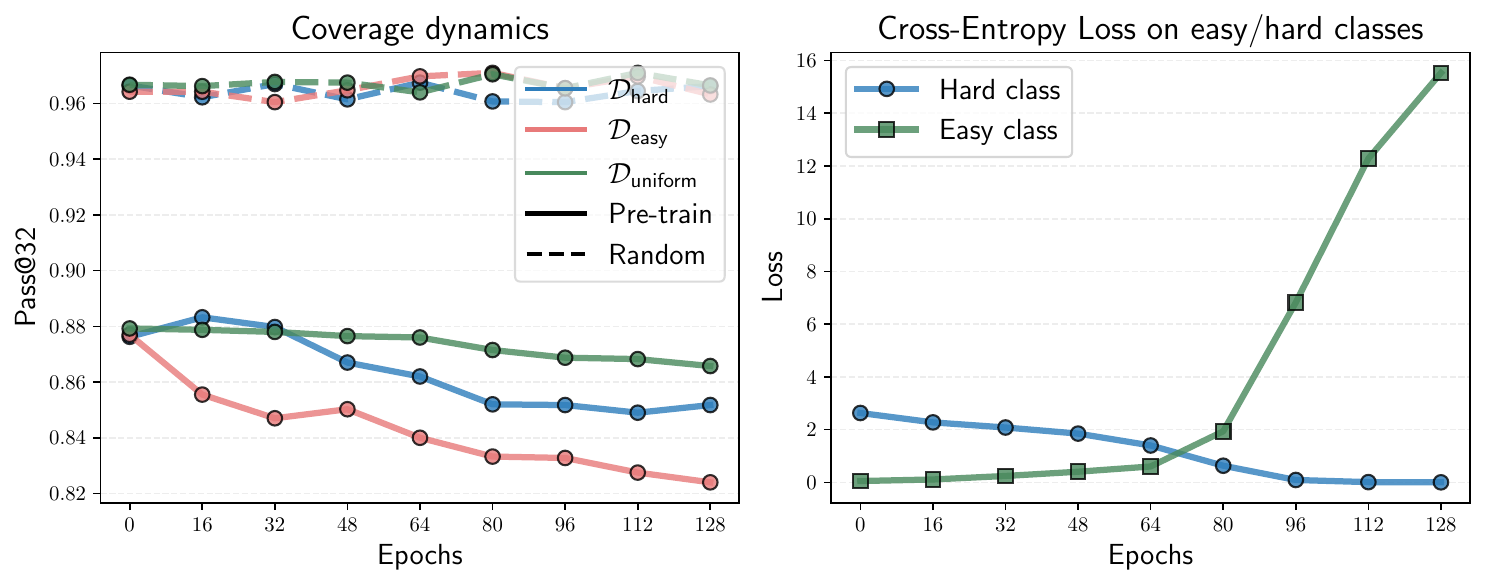}
    \vspace{-5pt}
    \caption{Coverage (Pass@$k$) on test inputs over the course of RLVR initialized from a pre-trained or randomly initialized model variants. Coverage shrinkage happens under both $\mathcal D_{\text{uniform}}$ and $\mathcal D_{\text{easy}}$, whereas learning on $\mathcal D_{\text{hard}}$ initially improves Pass@$k$ but begins to decline with prolonged training (Left), due to catastrophic forgetting on majority classes (Right).}
    \vspace{-15pt}
\label{fig:mnist_results}
\end{figure}

\looseness=-1 In this section, we study the impact of primacy bias in RLVR in a toy image classification setting where we can control the full pre-training and post-training pipeline, as well as input difficulty.\
Using a standard image classification task like MNIST \cite{LeCun1998GradientbasedLA}, we first pre-train a neural network on a labeled dataset and then fine-tune it with GRPO \citep{shao2024deepseekmathpushinglimitsmathematical}, where the reward is 1 if the predicted class matches the ground-truth label and 0 otherwise.
In this setting, we control the input’s difficulty based on how often its class appears in the pre-training dataset. We model this effect using Zipf's law distribution. So, the probability of observing the $i$-th most frequent class is given by:
\begin{align}
    p_i \propto i^{-\beta}, \quad i\in\{1,\cdots C\}
\end{align}
\looseness=-1 where $C$ is the number of classes, and $\beta>0$ is a parameter to control the skewness of the data distribution. As $\beta$ increases, the long tail becomes more pronounced, making the model focus on learning features from dominant classes, which can be suboptimal for low-frequency classes \citep{pmlr-v202-francazi23a,kunstner2024heavytailed} (as shown in Figure~\ref{fig:mnist_demonstration} (left)).
To study the role of data distribution in RLVR, we also construct three fine-tuning datasets: $\mathcal{D}_{\text{easy}}$, $\mathcal{D}_{\text{hard}}$, and $\mathcal{D}_{\text{uniform}}$, corresponding to data distributions that emphasize dominant classes, rare classes, and a uniform distribution over all classes, as shown in Figure~\ref{fig:mnist_demonstration} (right).
Our first observation is that RLVR training from a pre-trained model (using $ \beta=1.2$) leads to coverage shrinkage, whereas RLVR training from a randomly initialized model does not (as shown in Figure~\ref{fig:mnist_results}, left). Notably, this observation holds across different finetuning datasets.

\looseness=-1 Secondly, when fine-tuning from a pre-trained model, using easy problems leads to a shrinkage in coverage, where the model primarily sharpens its performance on already well-learned classes, similar to observations in post-trained LLMs \citep{yue2025does}. On the other hand, training on hard problems can initially improve Pass@$k$ but eventually degrades with prolonged training. As shown in Fig.~\ref{fig:mnist_results} (right), the loss on easy classes steadily increases over the course of training. This indicates that focusing on difficult examples induces forgetting of easier ones, ultimately leading to a decline in Pass@$k$.

\begin{figure}[!h]
    \vspace{-10pt}
    \centering
    \includegraphics[width=0.95\linewidth]{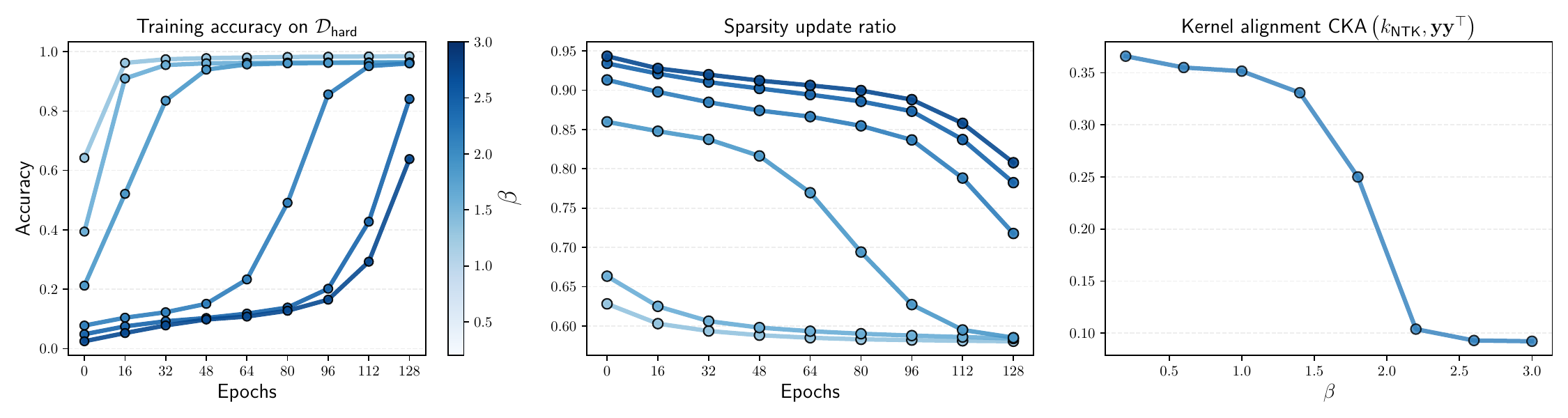}
    \vspace{-5pt}
    \caption{Skewness induces three key effects: it delays performance gains when learning on hard classes (left), results in sparser parameter updates during RLVR training (middle), and reduces alignment with the target labels on $\mathcal{D}_{\text{hard}}$. These results indicate that pre-trained models remain in a lazy regime during RLVR post-training, staying close to their initialization and requiring prolonged training to transition into an exploratory, feature-learning regime.}
    \label{fig:mnist_lazy}
    \vspace{-10 pt}
\end{figure}

\looseness=-1 Additionally, we examine the effect of skewness $\beta$ from pre-training to RL training. We find that larger skewness leads to: ($i$)~\emph{Delayed improvements on hard problems}, requiring substantially more training steps to reach perfect training accuracy on the hard examples (Fig.~\ref{fig:mnist_lazy}, left); and ($ii$)~\emph{Sparser parameter updates}, with only a small subset of parameters changing during RL training (Fig.~\ref{fig:mnist_lazy}, right).


Taken together, these findings provide evidence for a form of primacy bias in RLVR, whereby early training experiences hinder the model’s ability to learn rare classes or hard problems. Because the model relies heavily on pre-trained features when adapting to new data, it leads to limited parameter updates during RL, and delayed performance gains on hard examples. 

\vspace{-0.5em}
\subsection{Lazy learning regime}
\label{subsec:lazy_training}
\vspace{-0.5em}
Recently, \citet{Lyle2025WhatCG} have proposed a framework to understand primacy bias in deep learning. In this framework, primacy bias occurs because the network fails to enter a feature-learning regime after early training. Instead, the fine-tuning stage operates in a lazy regime where the network becomes increasingly rigid and does not produce substantial changes in the learned representation. When the data distribution or task changes, e.g., as in post-training, gradient updates are insufficient to significantly modify the previously learned features. 
As a result, the model continues to rely on representations optimized for the initial data rather than adapting to the new distribution, leading to degraded performance relative to a freshly initialized model.


Based on our observation and the above discussion, we hypothesize that RLVR is operated in the \emph{lazy regime}.
In this regime, the training dynamics stay close to the pre-trained initialization and a neural network $f_\theta(\x)$ can be well approximated by its first-order linearization around the initialization:
\begin{align}
    f_\theta(\x) \approx f_{\theta_0}(\x) + \nabla_\theta f(\x) \big|_{\theta_0} (\theta - \theta_0)
\end{align}
where $\theta_0$ denotes the pre-trained initialization of the model parameters. Furthermore, the learning dynamics are dictated by the initial neural tangent kernel (NTK) \citep{NEURIPS2018_5a4be1fa}: 
\begin{align}
    k_{\text{NTK}} = \left(\nabla_\theta f(\x) \big|_{\theta_0}\right) \left(\nabla_\theta f(\x) \big|_{\theta_0}\right)^\top
\end{align}

\paragraph{The learnability of different problem subsets.}

To identify which tasks the network can learn well in the lazy regime, several studies have proposed using centered kernel alignment, $\text{CKA}(k_{\text{NTK}}, \mathbf{y}\mathbf{y}^\top)$ \citep{Cortes2012AlgorithmsFL,Shan2021ATO}, as a measure of similarity between the initial NTK and the downstream target labels $\mathbf{y}$:
\begin{align}
    \text{CKA}\left(k_{\text{NTK}}, \y\y^\top\right) = \frac{\y^\top k_{\text{NTK}} \y}{||k_{\text{NTK}}||_F ||\y||^2}
\end{align}
A high CKA value indicates that the initial kernel is well-aligned with the target task, enabling good generalization with minimal feature learning, and generalizes poorly when the target labels $\y$ are \emph{misaligned} with the initial NTK. As shown in the third panel of Fig.~\ref{fig:mnist_lazy}, greater skewness leads to lower CKA alignment on $\mathcal{D}_{\text{hard}}$, indicating that delayed generalization can be understood as the misalignment of features.
In this transition, the model must adapt its internal representations to improve sample efficiency on functions that are poorly captured under the lazy regime \citep{Ghorbani2019LimitationsOL,kumar2024grokking,Arnaboldi2023FromH}.

\section{Fine-Tuning Reasoning Models with RLVR Occurs in the Lazy Regime}
\label{sec:linearized}
The above analysis suggests that primacy bias induced by pre-trained initialization drives the model into a lazy regime, limiting its ability to learn from hard problems. 
\looseness=-1 In this section, we investigate whether the \emph{lazy regime} persists during RLVR fine-tuning in reasoning models. To do so, we evaluate two criteria: (i)~whether the linearly approximated model closely matches the behavior of the fine-tuned model, and (ii)~whether RLVR training induces significant feature drift from the pre-trained initialization.
\subsection{RLVR models are well-approximated by linearization}
We compare the model and its corresponding linearized policy. Specifically, we linearly approximate the fine-tuned model logits, denoted as $\z_{\text{lin}}$, from the pre-trained model logits $\z_{\theta_0}$ as follows:
\begin{align}
\z_{\text{lin}}(\x) = \z_{\theta_0}(\x) + \nabla_{\theta} \z_{\theta}(\x)\big |_{\theta=\theta_0}^\top \left(\theta-\theta_0\right), \quad \pi_{\text{lin}}(\cdot | \x) = \mathrm{Softmax}\left( \z_{\text{lin}}(\x) \right).
\end{align}
where $\z_{\theta_0}$ denotes the logits of the pre-trained policy, to quantify the linearized approximation between the RL and SFT stages, as recent evidence suggests that SFT already operates in a near-lazy regime \citep{Malladi2022AKV, ren2025learning,afzal2026linearization}. We utilize the KL divergence between the fine-tuned models and their corresponding linearized approximations:
\begin{align}
    \mathbb E_{\x\sim\mathcal D}\left[\text{KL}\left(\pitheta(\cdot|\x) || \pi_{\text{lin}}(\cdot|\x)\right)\right],
\end{align}
We analyze publicly released checkpoints, as shown in Tab.~\ref{tab:linearized_kl}. The suite spans different RL algorithms (GRPO, DAPO) and domains (math, coding, instruction following).

\begin{table}[!ht]
\vspace{-10pt}
\centering
\caption{\small
\textbf{Linear approximation in SFT vs RLVR.} Linearized approximations reliably capture RL fine-tuned models, yet break down in the supervised fine-tuning (SFT) setting. 
}
\label{tab:linearized_kl}
\setlength{\tabcolsep}{4pt}
\resizebox{1.0\textwidth}{!}{%
\vspace{-10pt}
\begin{tabular}{@{} l l l l l r @{}}
\toprule
\textbf{Base Model} & \textbf{fine-tuned (FT) Model} & \textbf{Algorithm} & \textbf{Data} & $\textbf{KL}\left(\pitheta||\pi_{\text{lin}}\right)$ & \textbf{Sparsity Update (\%)}\\
\midrule
\href{https://huggingface.co/Qwen/Qwen2.5-Math-1.5B}{Qwen2.5-Math-1.5B} & 
\href{https://huggingface.co/deepseek-ai/DeepSeek-R1-Distill-Qwen-1.5B}{DS-R1-Distill-Qwen-1.5B} &
SFT & Mixed &  12.73 & 2.8\% \\
\href{https://huggingface.co/deepseek-ai/DeepSeek-R1-Distill-Qwen-1.5B}{DS-R1-Distill-Qwen-1.5B} &
\href{https://huggingface.co/agentica-org/DeepScaleR-1.5B-Preview}{DeepScaleR-1.5B-Preview} &
GRPO & Math & 0.06 & 53.8\% \\
\href{https://huggingface.co/deepseek-ai/DeepSeek-R1-Distill-Qwen-1.5B}{DS-R1-Distill-Qwen-1.5B} &
\href{https://huggingface.co/agentica-org/DeepCoder-1.5B-Preview}{DeepCoder-1.5B-Preview} &
GRPO & Code & 0.125 & 45.5\% \\
\midrule
\href{https://huggingface.co/Qwen/Qwen3-8B-Base}{Qwen3-8B-Base} &
\href{https://huggingface.co/Kwai-Klear/Klear-Reasoner-8B-SFT}{Klear-Reasoner-8B-SFT} &
SFT & Math+Code & 17.946 & 69.5\% \\
\href{https://huggingface.co/Qwen/Qwen3-8B-Base}{Qwen3-8B-Base} &
\href{https://huggingface.co/TMLR-Group-HF/GT-Qwen3-8B-Base}{GT-Qwen3-8B-Base} &
GRPO & Math & 0.042 & 79.9\% \\
\midrule
\href{https://huggingface.co/Qwen/Qwen3-4B}{Qwen3-4B} &
\href{https://huggingface.co/POLARIS-Project/Polaris-4B-Preview}{Polaris-4B-Preview} &
DAPO & Math & 0.052 & 79.3\%\\
\bottomrule
\end{tabular}}
\vspace{-5pt}
\end{table}

\begin{figure}[!h]
    \centering
\includegraphics[width=0.95\textwidth]{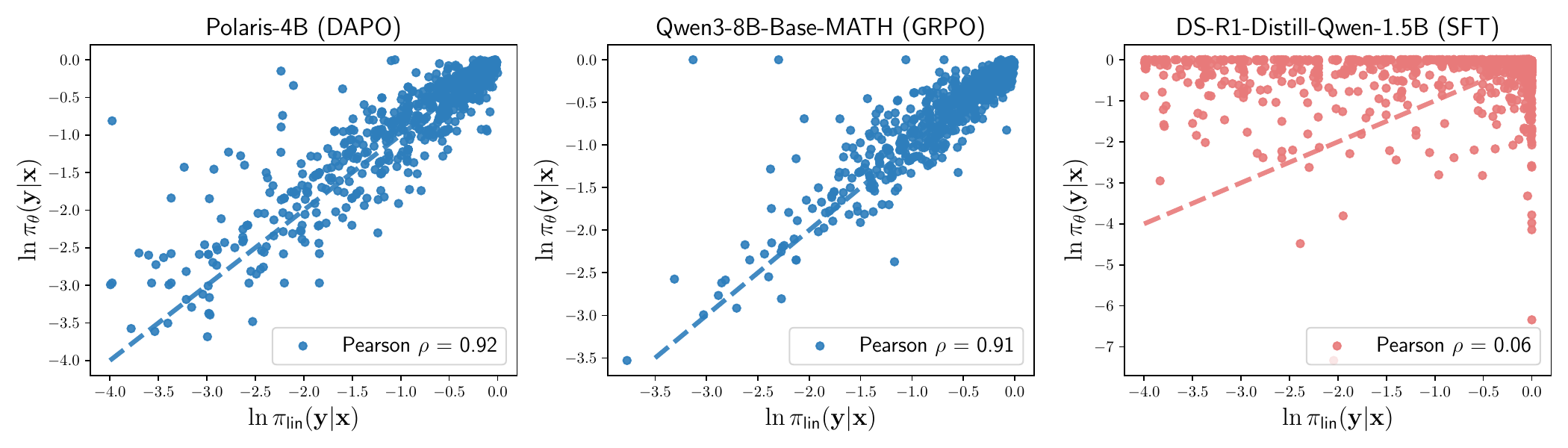}
    \vspace{-5pt}
\caption{Per-token correlations between the fine-tuned policy and its linearized approximation reveal that RLVR models align much more closely with their first-order approximations than SFT models.}
    \label{fig:linearized_corr}
    \vspace{-10pt}
\end{figure}

\looseness=-1 \textbf{Results.} As shown in Tab.~\ref{tab:linearized_kl}, interestingly, for all fine-tuned RL models, the KL divergence is extremely small, indicating that the linearized approximation closely matches the fully fine-tuned models. In contrast, supervised fine-tuning (SFT) yields much larger KL divergence, reflecting greater nonlinear changes. Moreover, RL-trained models exhibit a substantially higher update sparsity ratio than SFT, suggesting that they remain closer to their base model. We also quantify the per-token log-probability correlation between the linearized approximation and the final fine-tuned models. As shown in Fig.~\ref{fig:linearized_corr}, the linearized approximation closely matches the RL fine-tuned models, whereas the SFT models show little to no correlation.

\subsection{RLVR models show minimal feature drifts compared to SFT} 
\label{sec:feature_drift}
\looseness=-1 Another characteristic of the lazy–rich dichotomy in supervised learning \citep{keskar2017on,xing2019a} is that the lazy regime yields minimal representational change. To quantify feature similarity between pre-trained and fine-tuned models, we measure subspace rotation via the mean principal angle (MPA) \citep{chi-etal-2020-finding, zhou2025on} and spectral changes via spectrum drift (NSS). Specifically, given centered activation matrices $A^l_{\text{b}}$ and $A^l_{\text{ft}}$ at layer $l$, we compute MPA and NSS between them.
\begin{align}
    \text{MPA}(A^l_{\text{b}}, A^l_{\text{ft}}) &= \frac{1}{k}\sum\limits_{i=1}^k \text{arccos}\left(\sigma_i\left(V_{\text{b},k}(V_{\text{ft},k})^\top\right)\right), \text{ }   \text{NSS}(A^l_{\text{b}}, A^l_{\text{FT}}) = \frac{ ||\sigma(A^l_{\text{b}}) - \sigma(A^l_{\text{ft}}) ||_2}{||\sigma(A^l_{\text{b}})||_2}
\end{align}

\looseness=-1 where $V_{\text{b},k}V_{\text{ft},k} \in\mathbb R^{k\times d}$ denotes the top-$k$ right singular vectors of $A^l_{\text{b}}$, $A^l_{\text{ft}}$, and $\sigma_i$ denotes the singular value.

\begin{figure}[!h]
    \centering
    \vspace{-8pt}
    \includegraphics[width=0.95\linewidth]{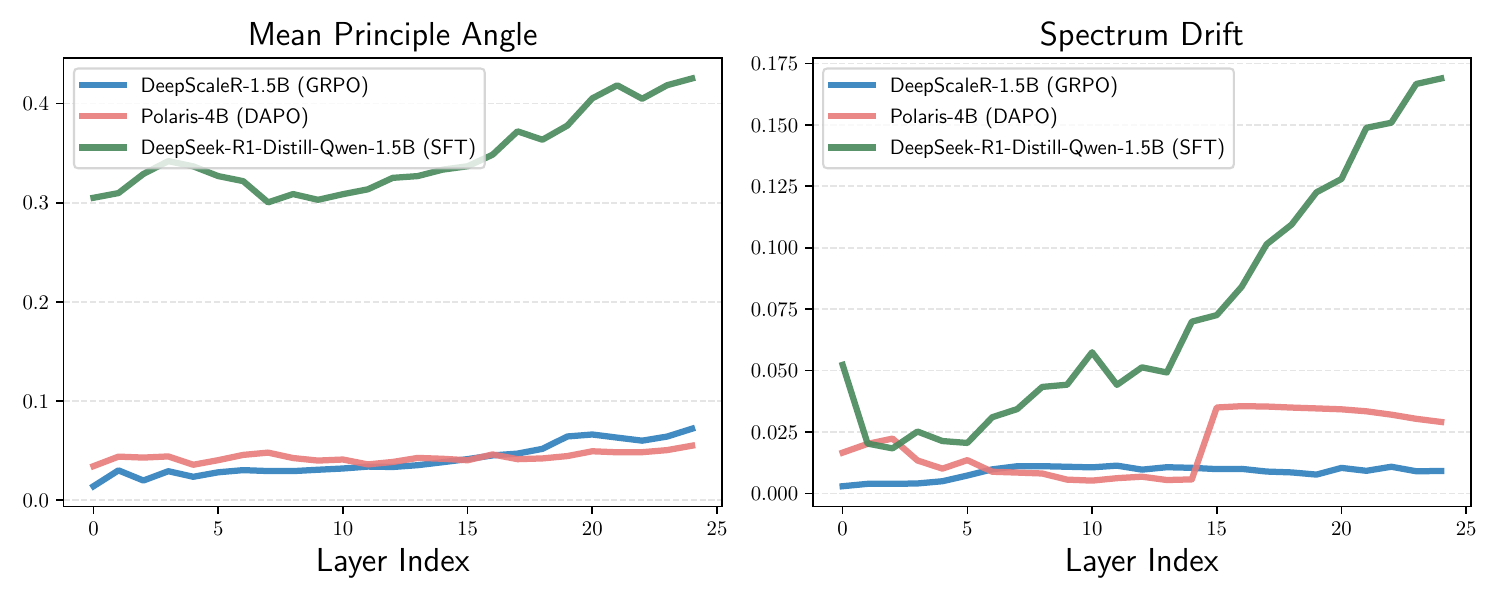}
    \vspace{-5pt}
    \caption{\looseness=-1 Mean Principal Angle (in radians) and Spectrum Drift (NSS) between the base and fine-tuned features across models. RL models show considerably lower shifts compared to SFT.}
    \vspace{-8pt}
    \label{fig:mpa_angle}
\end{figure}

\looseness=-1 The feature similarities between the pre-trained and their corresponding fine-tuned (SFT or RLVR) models across different layers are shown in Fig.~\ref{fig:mpa_angle}. Across models, RL fine-tuning yields \emph{consistently small} principal angles, indicating stronger preservation of the base model’s features. In contrast, SFT exhibits \emph{substantially larger} deviations from initial features.

To further support that LLM RL fine-tuning operates in the lazy regime, where the Neural Tangent Kernel (NTK) matrix mostly stays constant throughout the training process, 
we first calculate the empirical NTK (eNTK) on 256 unique samples across MLP layers using Qwen2.5-3B-Instruct, following previous work \cite{Malladi2022AKV}:
\begin{wrapfigure}{r}{0.45\textwidth}
\centering
\includegraphics[width=1.0\linewidth]{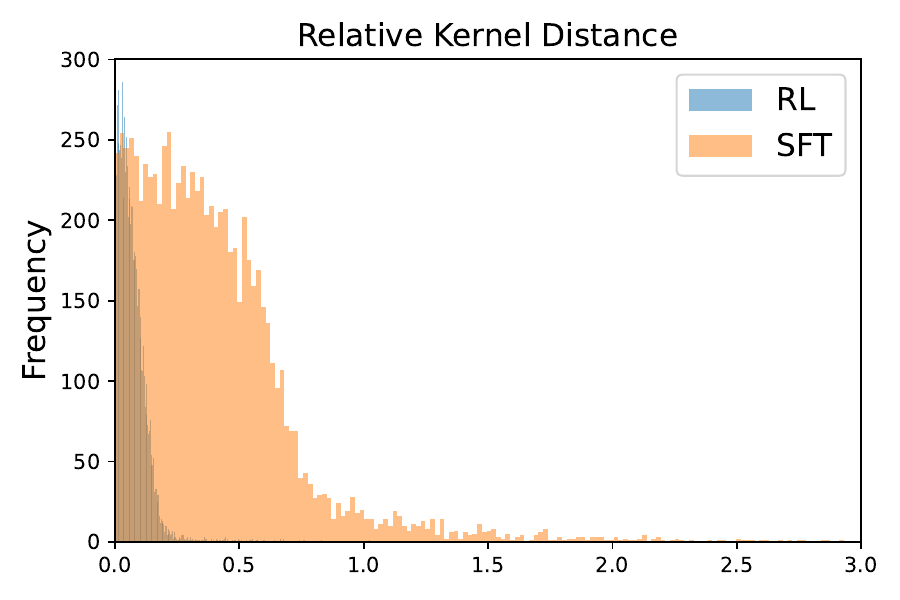}
     \vspace{-10pt}
\caption{Histogram of relative kernel distances between RL- and SFT-fine-tuned models across different examples.}
    \label{fig:relative_kernel_distance}
    \vspace{-10pt}
\end{wrapfigure}
\begin{align}
    \mathcal K^{\text{NTK}}(x_i,x_j) = \langle \nabla_\theta f_\theta (x_i), \nabla_\theta f_\theta(x_j)\rangle
\end{align}
where $f_\theta$ denotes the model predictions, we calculate the relative distance computed before and after fine-tuning:
\begin{align}
    \frac{ | \mathcal K^{\text{NTK}}_{\text{FT}}(x_i, x_j) - \mathcal K^{\text{NTK}}_{\text{base}}(x_i, x_j)|}{|\mathcal K^{\text{NTK}}_{\text{base}}(x_i, x_j)|}
\end{align}
where $\mathcal K^{\text{NTK}}_{\text{base}}$ denotes the eNTK of the base model and $\mathcal K^{\text{NTK}}_{\text{FT}}$ denotes the eNTK of the RL fine-tuned model. 

As shown in Fig.~\ref{fig:relative_kernel_distance}, RL remains tightly concentrated near zero with significantly less variance, indicating that the kernel changes only marginally during training. In contrast, SFT exhibits a much broader distribution that deviates substantially from zero, suggesting significantly larger kernel changes. These results suggest that RL fine-tuning operates in a \emph{lazy regime}, where the fine-tuned model is well-approximated by its linearization, and the learned representations remain largely unchanged throughout fine-tuning.


\section{The \looseness=-1 Unlearnable Phenomenon in RLVR Training.}
\label{sec:unlearnable}
In this section, we show that, due to primacy bias, LLMs primarily improve rewards on already known and easy problems throughout the RLVR training. As a result, they achieve only marginal gains on harder problems. We also observe that, while training on these hard problems yields little improvement, it can improve Pass@$k$ performance; however, prolonged training comes at the expense of reduced Pass@$1$ due to forgetting on easy problems.



\subsection{The harder the examples are, the harder it is for RLVR models to learn}

\looseness=-1 Our hypothesis is that, in the lazy regime, RLVR models rely on their initial representations to optimize rewards, which may be insufficient for harder examples and thus lead to degraded learning efficiency. 

\looseness=-1 \textbf{Setup.} We fine-tune the Qwen2.5-3B-Instruct model on the MATH dataset and filter out unsolved problems under the final policy to focus on solvable instances. The remaining data is partitioned into disjoint subsets of increasing difficulty based on pass@1 accuracy, and separate policies are trained on each subset to analyze how learning dynamics vary with difficulty level. 
Additional experimental details are provided in Appendix~\ref{app:exp_details}. In short, difficulty is closely related to CKA: well-aligned examples are easier to learn, whereas poorly aligned examples remain difficult. This not only supports our analysis above, but also sheds further light on the learning dynamics of RLVR on these examples. For evaluation, we utilize test sets from Olympiad-Bench \cite{he-etal-2024-olympiadbench}, Minerva \cite{xia2025minerva}, and MATH-500 \cite{hendrycks2021measuringmath500} datasets.

\begin{figure}[!h]
    \centering
    \includegraphics[width=0.95\linewidth]{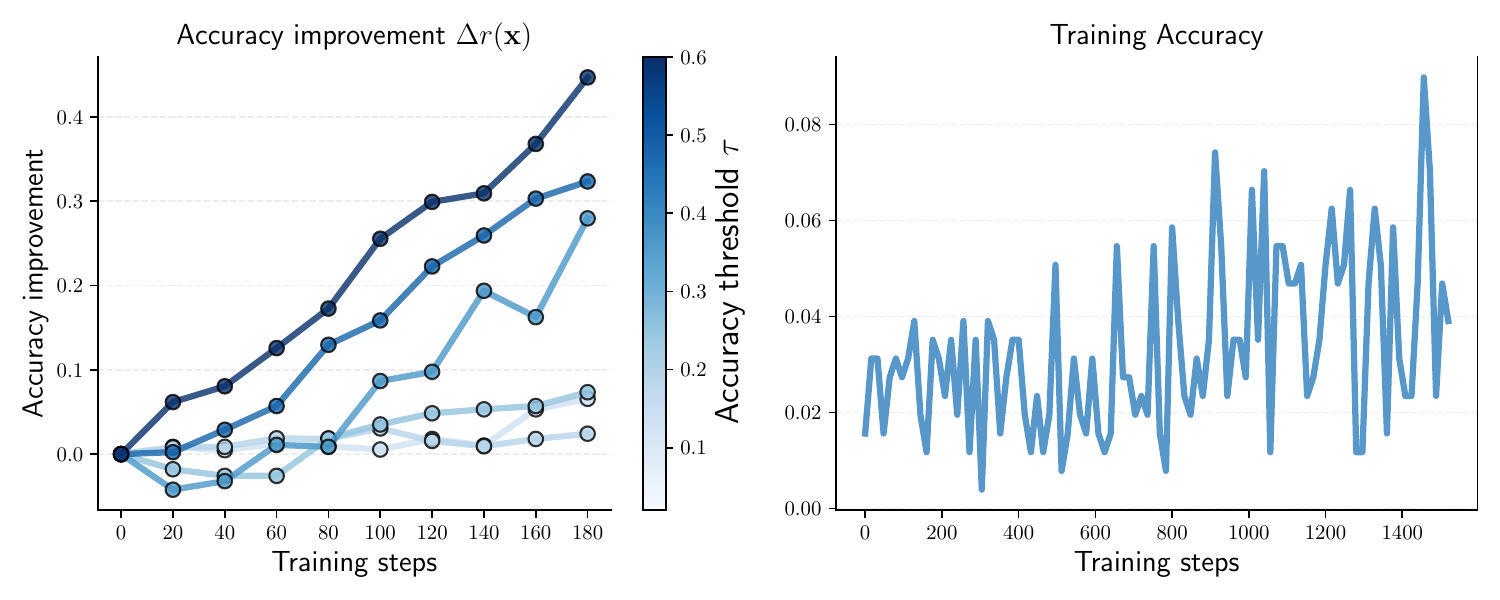}
    \vspace{-5pt}
    \caption{\looseness=-1 RLVR tends to improve only on easy subsets, while harder subsets exhibit minimal gains (left). Prolonged training on these hard subsets still yields limited improvement, failing to significantly enhance performance on training dataset.}
    \vspace{-5pt}
    \label{fig:reward_improvement}
\end{figure}
\looseness=-1 \textbf{Remark.} Throughout this paper, we define example difficulty by how challenging an example is for the model to solve, as measured by its accuracy. In the controlled toy experiments (Sec.~\ref{sec:primacy_bias_mnist}), we have full control over the pretraining data distribution, allowing us to manipulate difficulty by increasing distribution skew. In contrast, in the large-scale LLM experiments, where the pretraining data distribution is inaccessible, we cannot directly manipulate difficulty. Instead, we estimate it empirically using the model's accuracy on the MATH benchmark, with lower accuracy indicating greater difficulty. Despite these differences in how difficulty is manipulated or measured, both settings share the same underlying notion of difficulty.

\looseness=-1 \textbf{Training exclusively on hard examples yields little improvement.} Our first result is shown in Fig.~\ref{fig:reward_improvement}. We observe that RLVR yields meaningful improvements only on easy subsets, whereas on harder subsets, it shows little to no improvement, even when training is focused on them. Our results show that during training, most improvements arise from easier examples, while harder examples contribute minimally to overall reward gains. This provides further evidence to the phenomenon we observed earlier in Sec.~\ref{sec:primacy_bias_mnist}, where rare (hard) problems exhibit delayed improvement and require substantially more training steps due to a lack of feature learning, and prolonged training is needed to adapt their representations.

\begin{table}[ht]
\vspace{-10pt}
\centering
\small
\setlength{\tabcolsep}{10pt}  
\renewcommand{\arraystretch}{1.0}  
\caption{Comparison of Pass@128 performance when learning on hard/easy and full training dataset, where we train for 512 steps.}
\begin{tabular}{lccc}
\toprule
\multirow{1}{*}{\textbf{Model}} 
  & \textbf{OlympiadBench}
  & \textbf{Minerva}
  &\textbf{MATH-500} \\
\midrule
Base      
  & 68.77 (+0.0) & \textbf{53.91}(+0.0) &   94.31 (+0.0)
    \\
GRPO (full dataset)      
  & 67.87 (-0.9) & 48.83  (-5.08) & 93.24 (-1.07)
   \\
GRPO (Hard)             
  & \textbf{69.87} (+1.1) & 52.98 (-0.93) &  \textbf{94.51} (+0.2)
   \\
GRPO (Easy)       
  & 66.38 (-2.39)  & 49.18 (-4.73) & 92.47 (-1.84)
    \\
\bottomrule
\end{tabular}
\label{tab:data_comparison}
\vspace{-5pt}
\end{table}

\begin{figure}[!h]
    \centering
    \includegraphics[width=0.95\linewidth]{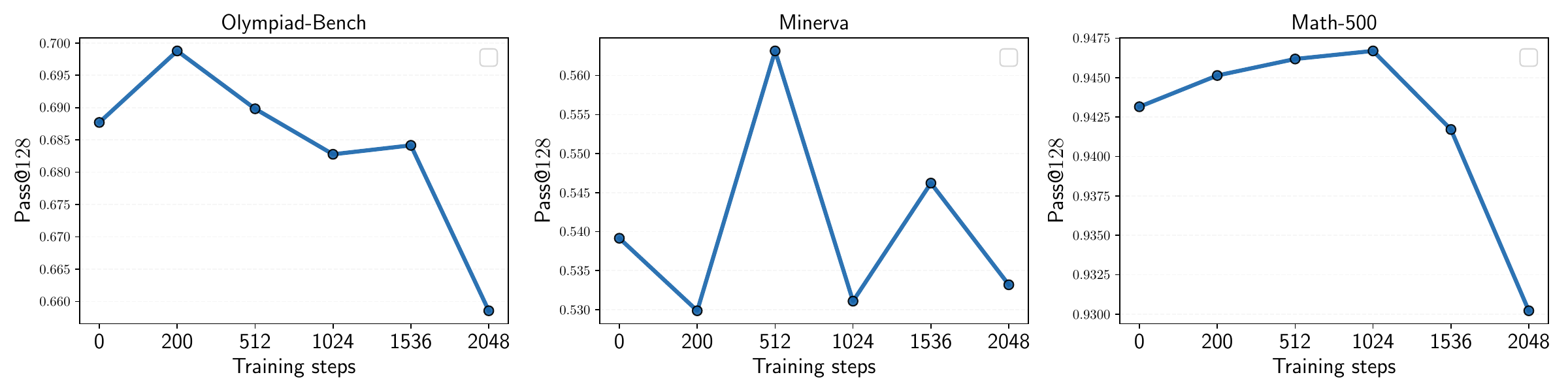}
    \vspace{-5pt}
    \caption{Pass@$k$ performance of the base model and intermediate checkpoints with GRPO (hard), when we prolonged training from 512 to 2048 training steps.}
    \label{fig:prolonged_passk}
\end{figure}
\begin{wrapfigure}{r}{0.45\textwidth}
\centering
\includegraphics[width=1.0\linewidth]{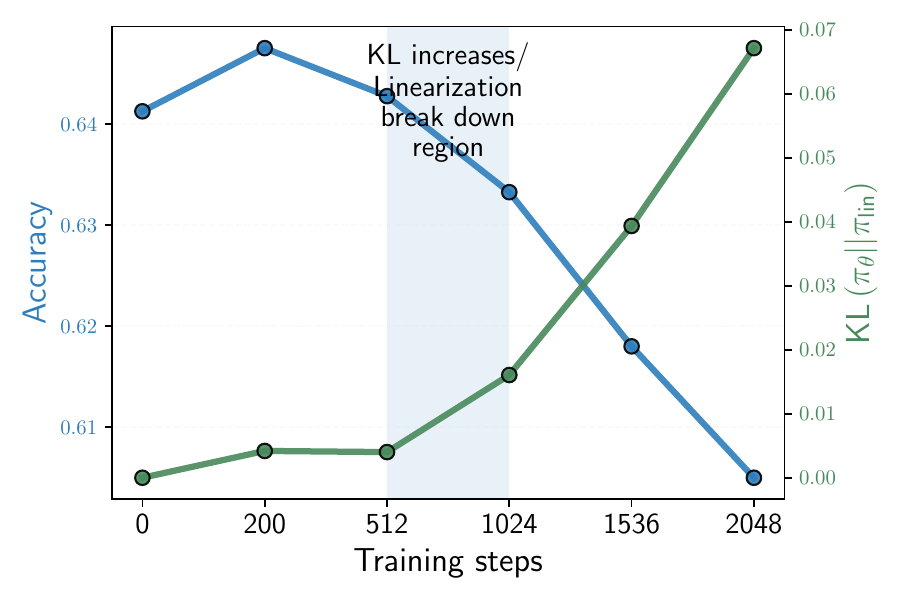}
     \vspace{-10pt}
\caption{RLVR models gradually forget easy examples (evaluated on MATH-500) when learning on hard problems, and the linearized approximation begins to break down under prolonged training.}
    \label{fig:prolonged_acc}
    \vspace{-10pt}
\end{wrapfigure}
\looseness=-1 \textbf{Training on hard examples improves coverage (Pass@$k$), while easy examples degrade it.} Interestingly, our evaluation results in Tab.~\ref{tab:data_comparison} show that despite limited improvement from training in hard subsets, the model still exhibits improved Pass@$k$, while learning on easy problems exhibits degradation in Pass@$k$ and coverage shrinkage. This suggests that these small gains can still transfer positively to other hard problems.

\textbf{Impact of prolonged training.} We further examine whether increasing gradient steps improves Pass@$k$, expecting the model to escape the lazy regime. Fig.~\ref{fig:prolonged_passk} shows its evolution over post-training steps. In contrast to \cite{pan2022directionalsharpness}, prolonged training actually degrades performance on the Pass@$k$ metric.
\looseness=-1 We hypothesize that this discrepancy stems from forgetting easier problems. As shown in Fig.~\ref{fig:prolonged_acc}, additional training steps tend to reduce average accuracy, suggesting that RLVR models gradually forget easy examples during training.

\begin{figure}[!h]
    \centering
    \vspace{-8pt}
    \includegraphics[width=0.95\linewidth]{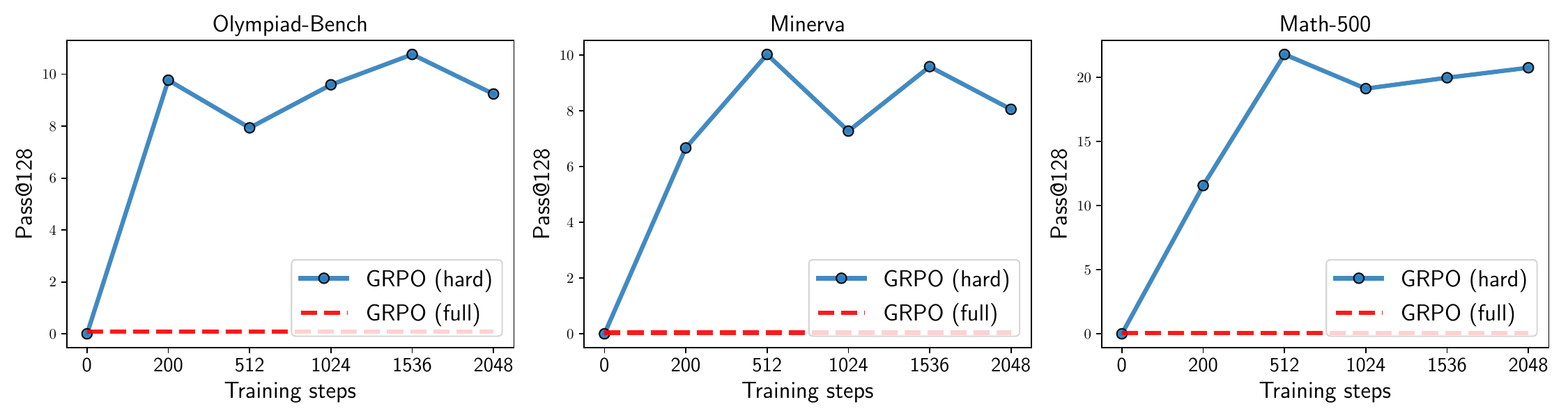}
    \vspace{-5pt}
    \caption{While Pass@$k$ (with $k=128$) degrades with prolonged training, this decline is primarily driven by forgetting on solvable problems; in contrast, the Pass@$k$ performance on problems that are unsolvable for the base model increases steadily when learning only on hard problems.}
    \label{fig:hard_acc}
    \vspace{-5pt}
\end{figure}

Pass@$k$ has been a standard metric for assessing the reasoning capabilities of LLMs. However, its value is inherently dependent on the specific evaluation set on which it is computed. In mathematical reasoning, Pass@$k$ is usually calculated over all test sets, including both already \emph{solvable} and \emph{unsolvable} problems. As a result, the decline in performance on already solvable problems might mask the improvement of RL-trained models on hard, unsolvable problems.
To quantify this phenomenon, we also compare with Pass@$k$ specifically computed on unsolvable problems (i.e., zero-accuracy problems) under the base model. As shown in Fig.~\ref{fig:hard_acc}, training on hard problems can significantly help models solve new problems and expand their coverage (Pass@$k$), in contrast to GRPO on the full dataset. This result further suggests that RLVR models can still acquire new capabilities under the prolonged training, but this gain comes at the cost of forgetting easier problems which might be hurtful in the simple well-trained tasks but mostly helpful if tasks are hard enough and new to models.

\textbf{Discussion.} Our analysis suggests that just exclusively learning on hard problems (without any algorithmic modification) could be sufficient to improve models' coverage (Pass@$k$). However, prolonged training can also degrade Pass@$k$ due to forgetting solvable/easy problems. This explains why, in mathematical reasoning — where Pass@$k$ is computed over a mixture of easy and hard problems — such forgetting can significantly hurt overall performance and hinder progress on hard problems. In contrast, on OOD and more difficult tasks \citep{liu2025prorl,sun2026rl}, this forgetting effect can actually be beneficial, since the base model performs poorly with initial representations and improvements on previously unsolved problems dominate the evaluation.
\section{Conclusion}
We investigate the ongoing debate over whether RLVR truly expands the reasoning boundary beyond what is acquired during pre-training, through the lens of primacy bias and the lazy training regime. Through extensive, controlled, and large-scale empirical analysis, we confirm that this bias emerges during RLVR training, causing models to remain confined around their pre-trained initialization and preventing meaningful adaptation to hard examples. We then uncover an \emph{unlearnable} phenomenon as a consequence of primacy bias, where RL favors improvement on easy problems, hindering the ability to learn harder ones. Consequently, attempting to escape the lazy regime can help models learn previously unsolvable problems, but at the cost of forgetting easier ones. This helps explain why, in OOD domains—where forgetting is less harmful—RL can expand the reasoning boundary, in contrast to mathematical reasoning.
\bibliographystyle{abbrvnat}
\bibliography{NeurIPS_2025/neurips2025_references}

\newpage
\appendix
\section{Limitations.}
\label{app:limitations}
This work primarily contributes to the ongoing debate on whether RLVR truly expands reasoning boundaries beyond those acquired during pre-training, attributing these conflicting views to the primacy bias phenomenon. It may help practitioners detect and mitigate the trade-off between adaptability and forgetting in post-training, thereby improving reliability in verifiable domains such as math and code. However, stronger RLVR training may also enhance general reasoning capabilities that could be misused. Moreover, methods relying on automated verifiers are susceptible to reward hacking and over-optimization, while diversity-preserving interventions may increase the likelihood of rare undesirable outputs. Our empirical analysis is limited to finite evaluation sets and task-specific reward signals. Extending our analysis to coding and other verifiable domains would be valuable. However, due to computational constraints, particularly the high cost of sandbox-based program execution for coding RLVR, we leave these extensions for future work.
\section{Additional Experimental Details.}
\label{app:exp_details}
\textbf{Models.} We utilize Qwen2.5-3B-Instruct \cite{qwen2025qwen25technicalreport} for RLVR training.

\textbf{Training configurations.} We utilize the Verl framework \citep{sheng2024hybridflowverl} for RLVR training. We train each model on 4 GPUs with a constant learning rate of $10^{-6}$, the maximum response length is 3072, with a total batch size of 256. For each prompt, we collect 16 rollouts to compute advantages for the GRPO update. We use a sampling temperature $\tau=1$. We do not apply Reverse KL
divergence or entropy loss in our training. We train each
model on 4 Nvidia H100 GPUs.

\textbf{Details constructing difficulty subset of training data.} We fine-tune the Qwen2.5-3B-Instruct model \citep{qwen2025qwen25technicalreport} on the MATH training dataset $\mathcal{D}$ \citep{hendrycks2021measuringmath500}, which consists of 7500 training examples, to identify difficult examples after RL fine-tuning. 
Full experimental details are provided in Appendix~\ref{app:exp_details}. 
After fine-tuning, we discard all problems with zero accuracy under the final policy $\pi_\theta$, retaining only solvable instances. 
We then partition $\mathcal{D}$ into $N$ \emph{disjoint} subsets $\{\mathcal{D}_k\}_{k=1}^N$ according to their estimated accuracy under $\pi_\theta$, ordered by increasing difficulty:
\begin{align}
\mathcal{D}_k = \{ \mathbf{x} \mid \tau_{k-1} < \text{pass@}1( \mathbf{x}) \leq \tau_k \}, \quad k \in [N],
\end{align}
where $\{\tau_k\}_{k=1}^N$ is an increasing sequence of thresholds defining the difficulty levels. We then fine-tune a separate policy on each subset $\mathcal{D}_k$ to study how learning dynamics vary with problem difficulty. To quantify improvements, we define the accuracy improvement metric before and after fine-tuning on these subsets:
\begin{align}
    \Delta r(\x) = \mathbb E_{\y\sim\pitheta(\cdot|\x)}[r(\x,\y)] - \mathbb E_{\y\sim\pi_b(\cdot|\x)}[r(\x,\y)] 
\end{align}
To estimate the average accuracy of the RL-trained policy, we sample $G=64$ responses per problem with a temperature of $1.0$. Finally, we construct $N$ disjoint subsets, where each subset contains problems whose scores fall within the interval $[\tau_k, \tau_{k+1}]$, for $k \in [N]$.
\newpage
\section{Gradient norm in RL-training.}
\begin{figure}[!h]
    \centering
    \includegraphics[width=0.8\linewidth]{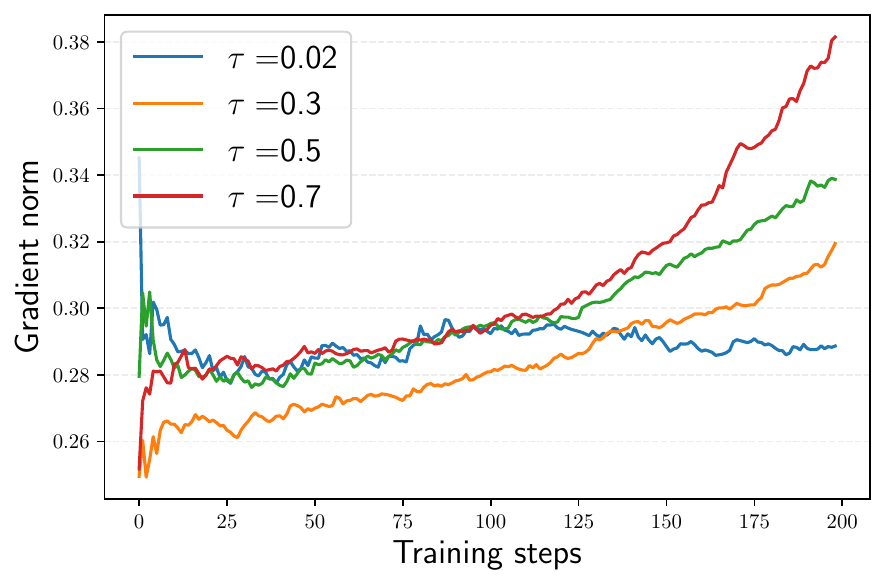}
    \vspace{-5pt}
    \caption{Gradient norm on different difficulty subsets; easy subsets tend to have larger gradient norm, indicating the RL models make faster progress on easy problems.}
    \label{fig:grad_norm_unlearn}
\end{figure}
In this section, we investigate the gradient norm when learning under easy/hard problems. We argue that the gradient is dominated by the easy problems, and, as a result is biased towards making progress on easy problems at the expense of hard problems. Fig.~\ref{fig:grad_norm_unlearn} demonstrates our results. Harder examples tend to have smaller gradient norms, suggesting that RLVR makes faster progress on easy problems compared to harder ones.
\section{Additional experiments on Llama-3.2-3B-Instruct.}
To strengthen the empirical evidence, we conducted additional experiments on Llama-3.2-3B-Instruct, using the same training setup as in our Qwen2.5-3B experiments. Specifically, we compare training on the full dataset against training exclusively on hard examples. 

\begin{figure}[!h]
\vspace{-10pt}
    \centering
    \includegraphics[width=0.6\linewidth]{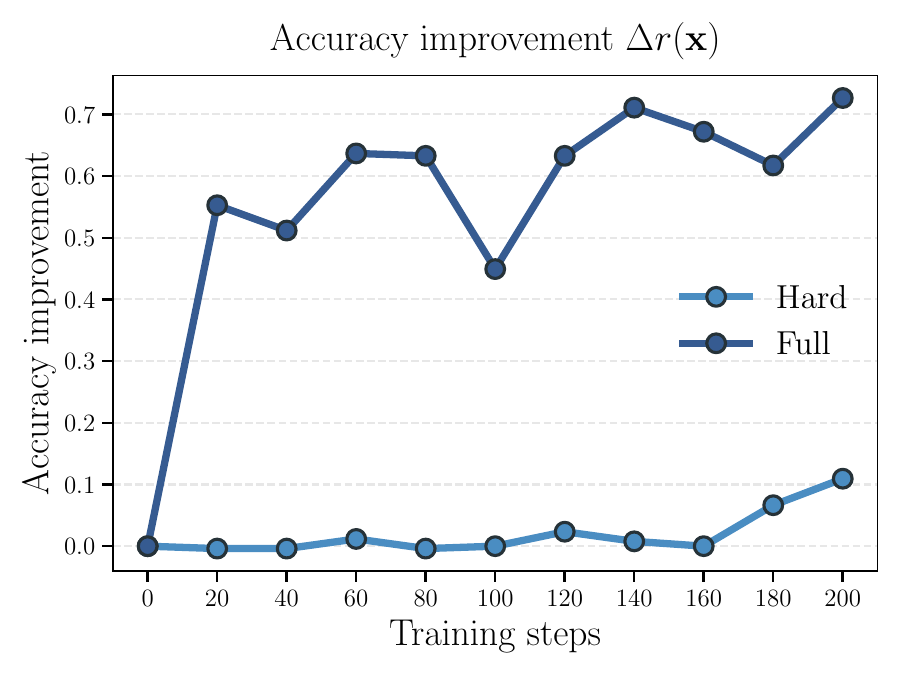}
    \vspace{-5pt}
    \caption{Similar to Qwen2.5-3B-Instruct, Llama-3.2-3B-Instruct exhibits rapid improvement when trained on the full dataset but shows minimal learning when trained on hard subsets.}
    \label{fig:llama_accuracy_improvement}
\end{figure}
We provide our results in Fig.~\ref{fig:llama_accuracy_improvement}, showing that training exclusively on hard examples leads to substantially slower optimization. Training on the full dataset rapidly improves training accuracy, 
\begin{table}[ht]
\vspace{-10pt}
\centering
\small
\setlength{\tabcolsep}{10pt}  
\renewcommand{\arraystretch}{1.0}  
\caption{Comparison of Pass@128 performance when learning on hard and full training datasets under Llama-3.2-3B-Instruct.}
\begin{tabular}{lccc}
\toprule
\multirow{1}{*}{\textbf{Model}} 
  & \textbf{OlympiadBench}
  & \textbf{Minerva}
  &\textbf{MATH-500} \\
\midrule
Base      
  & \textbf{61.01} & 51.37 &  \textbf{91.30}
    \\
GRPO (full dataset)      
  & 50.36 & 51.04 & 83.4 
   \\
GRPO (Hard)             
  & 	57.8 &	\textbf{51.61} & 89.1
   \\
\bottomrule
\end{tabular}
\label{tab:llama_data_comparison}
\end{table}
Furthermore, Table~\ref{tab:llama_data_comparison} shows that, on the new model backbone, hard-only training consistently achieves higher Pass@128 on problems unsolvable by the base model and maintains better Pass@$k$ performance than GRPO training on the full dataset, consistent with our earlier observations on Qwen2.5-3B-Instruct.
\newpage
\section*{NeurIPS Paper Checklist}

\begin{enumerate}

\item {\bf Claims}
    \item[] Question: Do the main claims made in the abstract and introduction accurately reflect the paper's contributions and scope?
    \item[] Answer: \answerYes{} 
    \item[] Justification: We uncover the primacy bias and lazy training regime in Large Language Models (Sections~\ref{sec:primacy_bias_mnist} and ~\ref{sec:linearized}), we further uncover the unlearable phenomenon during RLVR training, addressing the conflicting debate on whether RL can truly expands reasoning boundary (Section~\ref{sec:unlearnable}).
    \item[] Guidelines:
    \begin{itemize}
        \item The answer \answerNA{} means that the abstract and introduction do not include the claims made in the paper.
        \item The abstract and/or introduction should clearly state the claims made, including the contributions made in the paper and important assumptions and limitations. A \answerNo{} or \answerNA{} answer to this question will not be perceived well by the reviewers. 
        \item The claims made should match theoretical and experimental results, and reflect how much the results can be expected to generalize to other settings. 
        \item It is fine to include aspirational goals as motivation as long as it is clear that these goals are not attained by the paper. 
    \end{itemize}

\item {\bf Limitations}
    \item[] Question: Does the paper discuss the limitations of the work performed by the authors?
    \item[] Answer: \answerYes{} 
    \item[] Justification: We discuss the limitations of the work in Appendix~\ref{app:limitations}
    \item[] Guidelines:
    \begin{itemize}
        \item The answer \answerNA{} means that the paper has no limitation while the answer \answerNo{} means that the paper has limitations, but those are not discussed in the paper. 
        \item The authors are encouraged to create a separate ``Limitations'' section in their paper.
        \item The paper should point out any strong assumptions and how robust the results are to violations of these assumptions (e.g., independence assumptions, noiseless settings, model well-specification, asymptotic approximations only holding locally). The authors should reflect on how these assumptions might be violated in practice and what the implications would be.
        \item The authors should reflect on the scope of the claims made, e.g., if the approach was only tested on a few datasets or with a few runs. In general, empirical results often depend on implicit assumptions, which should be articulated.
        \item The authors should reflect on the factors that influence the performance of the approach. For example, a facial recognition algorithm may perform poorly when image resolution is low or images are taken in low lighting. Or a speech-to-text system might not be used reliably to provide closed captions for online lectures because it fails to handle technical jargon.
        \item The authors should discuss the computational efficiency of the proposed algorithms and how they scale with dataset size.
        \item If applicable, the authors should discuss possible limitations of their approach to address problems of privacy and fairness.
        \item While the authors might fear that complete honesty about limitations might be used by reviewers as grounds for rejection, a worse outcome might be that reviewers discover limitations that aren't acknowledged in the paper. The authors should use their best judgment and recognize that individual actions in favor of transparency play an important role in developing norms that preserve the integrity of the community. Reviewers will be specifically instructed to not penalize honesty concerning limitations.
    \end{itemize}

\item {\bf Theory assumptions and proofs}
    \item[] Question: For each theoretical result, does the paper provide the full set of assumptions and a complete (and correct) proof?
    \item[] Answer: \answerYes{} 
    \item[] Justification: We present our lazy regime and primacy bias assumption in Section~\ref {subsec:lazy_training}.
    \item[] Guidelines:
    \begin{itemize}
        \item The answer \answerNA{} means that the paper does not include theoretical results. 
        \item All the theorems, formulas, and proofs in the paper should be numbered and cross-referenced.
        \item All assumptions should be clearly stated or referenced in the statement of any theorems.
        \item The proofs can either appear in the main paper or the supplemental material, but if they appear in the supplemental material, the authors are encouraged to provide a short proof sketch to provide intuition. 
        \item Inversely, any informal proof provided in the core of the paper should be complemented by formal proofs provided in appendix or supplemental material.
        \item Theorems and Lemmas that the proof relies upon should be properly referenced. 
    \end{itemize}

    \item {\bf Experimental result reproducibility}
    \item[] Question: Does the paper fully disclose all the information needed to reproduce the main experimental results of the paper to the extent that it affects the main claims and/or conclusions of the paper (regardless of whether the code and data are provided or not)?
    \item[] Answer: \answerYes{}{} 
    \item[] Justification:  All implementation details are in \ref{app:exp_details}.
    \item[] Guidelines:
    \begin{itemize}
        \item The answer \answerNA{} means that the paper does not include experiments.
        \item If the paper includes experiments, a \answerNo{} answer to this question will not be perceived well by the reviewers: Making the paper reproducible is important, regardless of whether the code and data are provided or not.
        \item If the contribution is a dataset and\slash or model, the authors should describe the steps taken to make their results reproducible or verifiable. 
        \item Depending on the contribution, reproducibility can be accomplished in various ways. For example, if the contribution is a novel architecture, describing the architecture fully might suffice, or if the contribution is a specific model and empirical evaluation, it may be necessary to either make it possible for others to replicate the model with the same dataset, or provide access to the model. In general. releasing code and data is often one good way to accomplish this, but reproducibility can also be provided via detailed instructions for how to replicate the results, access to a hosted model (e.g., in the case of a large language model), releasing of a model checkpoint, or other means that are appropriate to the research performed.
        \item While NeurIPS does not require releasing code, the conference does require all submissions to provide some reasonable avenue for reproducibility, which may depend on the nature of the contribution. For example
        \begin{enumerate}
            \item If the contribution is primarily a new algorithm, the paper should make it clear how to reproduce that algorithm.
            \item If the contribution is primarily a new model architecture, the paper should describe the architecture clearly and fully.
            \item If the contribution is a new model (e.g., a large language model), then there should either be a way to access this model for reproducing the results or a way to reproduce the model (e.g., with an open-source dataset or instructions for how to construct the dataset).
            \item We recognize that reproducibility may be tricky in some cases, in which case authors are welcome to describe the particular way they provide for reproducibility. In the case of closed-source models, it may be that access to the model is limited in some way (e.g., to registered users), but it should be possible for other researchers to have some path to reproducing or verifying the results.
        \end{enumerate}
    \end{itemize}

\item {\bf Open access to data and code}
    \item[] Question: Does the paper provide open access to the data and code, with sufficient instructions to faithfully reproduce the main experimental results, as described in supplemental material?
    \item[] Answer: \answerYes{} 
    \item[] Justification: The data and models are publicly available, and the code is available at \url{https://anonymous.4open.science/r/neurips26_sub-0DE8/}.
    \item[] Guidelines:
    \begin{itemize}
        \item The answer \answerNA{} means that paper does not include experiments requiring code.
        \item Please see the NeurIPS code and data submission guidelines (\url{https://neurips.cc/public/guides/CodeSubmissionPolicy}) for more details.
        \item While we encourage the release of code and data, we understand that this might not be possible, so \answerNo{} is an acceptable answer. Papers cannot be rejected simply for not including code, unless this is central to the contribution (e.g., for a new open-source benchmark).
        \item The instructions should contain the exact command and environment needed to run to reproduce the results. See the NeurIPS code and data submission guidelines (\url{https://neurips.cc/public/guides/CodeSubmissionPolicy}) for more details.
        \item The authors should provide instructions on data access and preparation, including how to access the raw data, preprocessed data, intermediate data, and generated data, etc.
        \item The authors should provide scripts to reproduce all experimental results for the new proposed method and baselines. If only a subset of experiments are reproducible, they should state which ones are omitted from the script and why.
        \item At submission time, to preserve anonymity, the authors should release anonymized versions (if applicable).
        \item Providing as much information as possible in supplemental material (appended to the paper) is recommended, but including URLs to data and code is permitted.
    \end{itemize}

\item {\bf Experimental setting/details}
    \item[] Question: Does the paper specify all the training and test details (e.g., data splits, hyperparameters, how they were chosen, type of optimizer) necessary to understand the results?
    \item[] Answer: \answerYes{} 
    \item[] Justification: All implementation details are in the Appendix ~\ref{app:exp_details}.
    \item[] Guidelines:
    \begin{itemize}
        \item The answer \answerNA{} means that the paper does not include experiments.
        \item The experimental setting should be presented in the core of the paper to a level of detail that is necessary to appreciate the results and make sense of them.
        \item The full details can be provided either with the code, in appendix, or as supplemental material.
    \end{itemize}

\item {\bf Experiment statistical significance}
    \item[] Question: Does the paper report error bars suitably and correctly defined or other appropriate information about the statistical significance of the experiments?
    \item[] Answer: \answerNo{} 
    \item[] Justification: Training Large Language models is time-consuming and compute-intensive.
Our experiments do not run multiple seeds on one configuration due to limited computing and financial budget. Our evaluation protocol is
similar to prior influential works in LLM Reasoning \citep{zhang2025interplaypretrainingmidtrainingrl,sun2026rl,yuan2026from}.
    \item[] Guidelines:
    \begin{itemize}
        \item The answer \answerNA{} means that the paper does not include experiments.
        \item The authors should answer \answerYes{} if the results are accompanied by error bars, confidence intervals, or statistical significance tests, at least for the experiments that support the main claims of the paper.
        \item The factors of variability that the error bars are capturing should be clearly stated (for example, train/test split, initialization, random drawing of some parameter, or overall run with given experimental conditions).
        \item The method for calculating the error bars should be explained (closed form formula, call to a library function, bootstrap, etc.)
        \item The assumptions made should be given (e.g., Normally distributed errors).
        \item It should be clear whether the error bar is the standard deviation or the standard error of the mean.
        \item It is OK to report 1-sigma error bars, but one should state it. The authors should preferably report a 2-sigma error bar than state that they have a 96\% CI, if the hypothesis of Normality of errors is not verified.
        \item For asymmetric distributions, the authors should be careful not to show in tables or figures symmetric error bars that would yield results that are out of range (e.g., negative error rates).
        \item If error bars are reported in tables or plots, the authors should explain in the text how they were calculated and reference the corresponding figures or tables in the text.
    \end{itemize}

\item {\bf Experiments compute resources}
    \item[] Question: For each experiment, does the paper provide sufficient information on the computer resources (type of compute workers, memory, time of execution) needed to reproduce the experiments?
    \item[] Answer: \answerYes{} 
    \item[] Justification: They are mentioned in Appendix~\ref{app:exp_details}.
    \item[] Guidelines:
    \begin{itemize}
        \item The answer \answerNA{} means that the paper does not include experiments.
        \item The paper should indicate the type of compute workers CPU or GPU, internal cluster, or cloud provider, including relevant memory and storage.
        \item The paper should provide the amount of compute required for each of the individual experimental runs as well as estimate the total compute. 
        \item The paper should disclose whether the full research project required more compute than the experiments reported in the paper (e.g., preliminary or failed experiments that didn't make it into the paper). 
    \end{itemize}
    
\item {\bf Code of ethics}
    \item[] Question: Does the research conducted in the paper conform, in every respect, with the NeurIPS Code of Ethics \url{https://neurips.cc/public/EthicsGuidelines}?
    \item[] Answer: \answerYes{} 
    \item[] Justification: This work solely focuses on understanding the ongoing debate of coverage expansion/shrinkage problem in RLVR training. No human subject or private data is involved in this work, and no experiment is conducted for harmful purposes.
    \item[] Guidelines:
    \begin{itemize}
        \item The answer \answerNA{} means that the authors have not reviewed the NeurIPS Code of Ethics.
        \item If the authors answer \answerNo, they should explain the special circumstances that require a deviation from the Code of Ethics.
        \item The authors should make sure to preserve anonymity (e.g., if there is a special consideration due to laws or regulations in their jurisdiction).
    \end{itemize}

\item {\bf Broader impacts}
    \item[] Question: Does the paper discuss both potential positive societal impacts and negative societal impacts of the work performed?
    \item[] Answer: \answerYes{} 
    \item[] Justification: We mentioned in the Appendix ~\ref{app:limitations}.
    \item[] Guidelines:
    \begin{itemize}
        \item The answer \answerNA{} means that there is no societal impact of the work performed.
        \item If the authors answer \answerNA{} or \answerNo, they should explain why their work has no societal impact or why the paper does not address societal impact.
        \item Examples of negative societal impacts include potential malicious or unintended uses (e.g., disinformation, generating fake profiles, surveillance), fairness considerations (e.g., deployment of technologies that could make decisions that unfairly impact specific groups), privacy considerations, and security considerations.
        \item The conference expects that many papers will be foundational research and not tied to particular applications, let alone deployments. However, if there is a direct path to any negative applications, the authors should point it out. For example, it is legitimate to point out that an improvement in the quality of generative models could be used to generate Deepfakes for disinformation. On the other hand, it is not needed to point out that a generic algorithm for optimizing neural networks could enable people to train models that generate Deepfakes faster.
        \item The authors should consider possible harms that could arise when the technology is being used as intended and functioning correctly, harms that could arise when the technology is being used as intended but gives incorrect results, and harms following from (intentional or unintentional) misuse of the technology.
        \item If there are negative societal impacts, the authors could also discuss possible mitigation strategies (e.g., gated release of models, providing defenses in addition to attacks, mechanisms for monitoring misuse, mechanisms to monitor how a system learns from feedback over time, improving the efficiency and accessibility of ML).
    \end{itemize}
    
\item {\bf Safeguards}
    \item[] Question: Does the paper describe safeguards that have been put in place for responsible release of data or models that have a high risk for misuse (e.g., pre-trained language models, image generators, or scraped datasets)?
    \item[] Answer: \answerNo{} 
    \item[] Justification: This work solely focuses on understanding the ongoing debate between the coverage expansion/shrinkage problem in RLVR training. No new data or pretrained model is released in this work.
    \item[] Guidelines:
    \begin{itemize}
        \item The answer \answerNA{} means that the paper poses no such risks.
        \item Released models that have a high risk for misuse or dual-use should be released with necessary safeguards to allow for controlled use of the model, for example by requiring that users adhere to usage guidelines or restrictions to access the model or implementing safety filters. 
        \item Datasets that have been scraped from the Internet could pose safety risks. The authors should describe how they avoided releasing unsafe images.
        \item We recognize that providing effective safeguards is challenging, and many papers do not require this, but we encourage authors to take this into account and make a best faith effort.
    \end{itemize}

\item {\bf Licenses for existing assets}
    \item[] Question: Are the creators or original owners of assets (e.g., code, data, models), used in the paper, properly credited and are the license and terms of use explicitly mentioned and properly respected?
    \item[] Answer: \answerYes{}{} 
    \item[] Justification: All data used in this work are publicly available for research purposes and are cited accordingly.
    \item[] Guidelines:
    \begin{itemize}
        \item The answer \answerNA{} means that the paper does not use existing assets.
        \item The authors should cite the original paper that produced the code package or dataset.
        \item The authors should state which version of the asset is used and, if possible, include a URL.
        \item The name of the license (e.g., CC-BY 4.0) should be included for each asset.
        \item For scraped data from a particular source (e.g., website), the copyright and terms of service of that source should be provided.
        \item If assets are released, the license, copyright information, and terms of use in the package should be provided. For popular datasets, \url{paperswithcode.com/datasets} has curated licenses for some datasets. Their licensing guide can help determine the license of a dataset.
        \item For existing datasets that are re-packaged, both the original license and the license of the derived asset (if it has changed) should be provided.
        \item If this information is not available online, the authors are encouraged to reach out to the asset's creators.
    \end{itemize}

\item {\bf New assets}
    \item[] Question: Are new assets introduced in the paper well documented and is the documentation provided alongside the assets?
    \item[] Answer: \answerNA{}{} 
    \item[] Justification: This work solely focuses on understanding the ongoing debate between the coverage expansion/shrinkage problem in RLVR training. No new data or pretrained model is released in this work.
    \item[] Guidelines:
    \begin{itemize}
        \item The answer \answerNA{} means that the paper does not release new assets.
        \item Researchers should communicate the details of the dataset\slash code\slash model as part of their submissions via structured templates. This includes details about training, license, limitations, etc. 
        \item The paper should discuss whether and how consent was obtained from people whose asset is used.
        \item At submission time, remember to anonymize your assets (if applicable). You can either create an anonymized URL or include an anonymized zip file.
    \end{itemize}

\item {\bf Crowdsourcing and research with human subjects}
    \item[] Question: For crowdsourcing experiments and research with human subjects, does the paper include the full text of instructions given to participants and screenshots, if applicable, as well as details about compensation (if any)? 
    \item[] Answer: \answerNA{}{} 
    \item[] Justification: This work solely focuses on understanding the ongoing debate between the
coverage expansion/shrinkage problem in RLVR training. No human subject is involved in this work.
    \item[] Guidelines:
    \begin{itemize}
        \item The answer \answerNA{} means that the paper does not involve crowdsourcing nor research with human subjects.
        \item Including this information in the supplemental material is fine, but if the main contribution of the paper involves human subjects, then as much detail as possible should be included in the main paper. 
        \item According to the NeurIPS Code of Ethics, workers involved in data collection, curation, or other labor should be paid at least the minimum wage in the country of the data collector. 
    \end{itemize}

\item {\bf Institutional review board (IRB) approvals or equivalent for research with human subjects}
    \item[] Question: Does the paper describe potential risks incurred by study participants, whether such risks were disclosed to the subjects, and whether Institutional Review Board (IRB) approvals (or an equivalent approval/review based on the requirements of your country or institution) were obtained?
    \item[] Answer: \answerNA{} 
    \item[] Justification: This work solely focuses on understanding the ongoing debate between the
coverage expansion/shrinkage problem in RLVR training. No human subject is involved in this work.
    \item[] Guidelines:
    \begin{itemize}
        \item The answer \answerNA{} means that the paper does not involve crowdsourcing nor research with human subjects.
        \item Depending on the country in which research is conducted, IRB approval (or equivalent) may be required for any human subjects research. If you obtained IRB approval, you should clearly state this in the paper. 
        \item We recognize that the procedures for this may vary significantly between institutions and locations, and we expect authors to adhere to the NeurIPS Code of Ethics and the guidelines for their institution. 
        \item For initial submissions, do not include any information that would break anonymity (if applicable), such as the institution conducting the review.
    \end{itemize}

\item {\bf Declaration of LLM usage}
    \item[] Question: Does the paper describe the usage of LLMs if it is an important, original, or non-standard component of the core methods in this research? Note that if the LLM is used only for writing, editing, or formatting purposes and does \emph{not} impact the core methodology, scientific rigor, or originality of the research, declaration is not required.
    \item[] Answer: \answerNA{} 
    \item[] Justification: We solely use LLM to help with writing, all core methods and experiments
1170 were proposed without LLM involvement.
    \item[] Guidelines:
    \begin{itemize}
        \item The answer \answerNA{} means that the core method development in this research does not involve LLMs as any important, original, or non-standard components.
        \item Please refer to our LLM policy in the NeurIPS handbook for what should or should not be described.
    \end{itemize}

\end{enumerate}
\end{document}